# Classification of Safety Events at Nuclear Sites using Large Language Models (LLMs)

**Mishca de Costa[1,2], Muhammad Anwar[1,2], Daniel Lau[1,2], and Issam Hammad[2]**

[1] Enterprise Digital Technology, Digital Technology and Services, Ontario Power Generation, Pickering, Ontario, Canada

[2] Department of Engineering Mathematics and Internetworking, Faculty of Engineering, Dalhousie University, Halifax, Nova Scotia

mishca.decosta@opg.com; muhammad.anwar@opg.com; daniel.lau@opg.com; issam.hammad@dal.ca

**Abstract**

This paper proposes the development of a Large Language Model (LLM)-based machine learning classifier designed to categorize Station Condition Records (SCRs) at nuclear power stations into safety-related and non-safety-related categories. The primary objective is to augment the existing manual review process by enhancing the efficiency and accuracy of the safety classification process at nuclear stations. The paper discusses experiments performed to classify a labeled SCR dataset and evaluates the performance of the classifier. It explores the construction of several prompt variations and their observed effects on the LLM's decision-making process. Additionally, it introduces a numerical scoring mechanism that could offer a more nuanced and flexible approach to SCR safety classification. This method represents an innovative step in nuclear safety management, providing a scalable tool for the identification of safety events.

## 1.   Introduction

Station Condition Records (SCRs) are reports documenting events or conditions that occur at a nuclear power generating station, as well as initial actions taken or planned in response. An SCR that is assessed as relevant to safety goes through extra scrutiny to maintain personnel safety at the nuclear station. The current method of SCR classification is a manual one that involves human evaluators to examine multiple SCRs every week. These records, which may be submitted by any employee, cover a broad spectrum of events and undergo management review to determine an appropriate reaction. If an SCR is deemed relevant to safety, it undergoes further examination by the Health and Safety department and is documented in a specialized database. The SCR database encompasses a range of occurrences, from equipment malfunctions and delays in material delivery to staff missing training sessions, making it cumbersome for the Health and Safety department to sift through each SCR to identify safety-related items before transferring pertinent details into their safety tracking system. The aim of this project is to develop a machine learning classifier to automatically differentiate between safety-related and non-safety-related SCRs. While this tool is not intended to supplant human assessment, it will serve as an additional layer of scrutiny and facilitate the swift review of safety-related SCRs by triggering a pipeline that copies all relevant data into the safety system for final human verification.

This paper employs GPT-4, a large language model from OpenAI built on transformer architecture, to perform this classification. Although SCRs often contain domain-specific information that may challenge GPT-4, records concerning safety events typically exclude such specialized language. For instance, while a record might describe an incident where an employee was exposed to a hazardous substance during an ion exchange resin swap, it does not require an understanding of the ion exchange process itself for GPT-4 to recognize the exposure to a hazard.

2.      Literature Review

Utilizing AI for CANDU-type nuclear power plants was previously proposed by several researchers to enhance safety and operational efficiencies. Budzinski's work [1] highlights machine learning techniques for the verification of refueling activities in CANDU-type nuclear power plants with direct applications in nuclear safeguards. Ahsan and Hassan [2] develop a machine learning-based fault prediction system for the primary heat transport system of CANDU type pressurized heavy water reactors, illustrating its potential to pre-emptively address system failures and enhance reactor safety. Hammad et al. [3-4] utilize deep learning and linear regression to automate the detection of flaws in nuclear fuel channel ultrasonic scans, significantly advancing the automation of safety inspections. Wallace et al. [5] improve the localization of fuel defects using neural networks and ancillary data sources, and in [6] developed an ultrasonic inspection analysis tool that supports decision-making in defect identification.

Focusing on Natural Language Processing, Kant et al. [7] utilize a Transformer model for sentiment classification which, despite achieving a moderate 0.69 F1 score, required substantial computational resources and time for training. The fine-tuning process involved intricate balancing and additional strategies like active learning to address label nuances and class imbalances. Zhao et al. [8] present an approach to interpret CNN-based text classification using SHAP values, providing a deeper understanding of model decisions but adding complexity to the CNN model. Abburi et al. [9] explore the use of ensemble methods with various pre-trained LLMs, including BERT and its variants, for text classification. These models, while not achieving top results in binary classification tasks, demonstrated potential for nuanced classification and model attribution. In summary, while CNNs and LSTMs require significant effort to train and are sensitive to data imbalances and nuances, the reviewed works suggest that LLMs, especially when used in ensemble configurations, can more naturally capture the subtleties of language, justifying their use over traditional machine learning models for complex text classification tasks without requiring significant compute time or costs when using foundational models like GPT-4.

3.      Exploratory Data Analysis

To tackle this problem, we collected a random sample of 100,000 SCRs from the Safety Information Database (SIDB). These SCRs were selected for inclusion in SIDB based on various search criteria and were subsequently reviewed and classified by humans as either safety or non-safety events.
This SIDB database therefore contains both safety-positive and -negative samples; in fact, the non-safety records outnumber the safety records in our sample by a ratio of approximately 9:1 (90358 non-safety-related SCRs versus 9642 safety-related records).

This data source was chosen because the records it contains were subjected to a more thorough review with respect to safety than records in the larger SCR database. This allows us to place greater confidence in the human-made labels which are used to train the machine-learning classifier.

With this highly imbalanced dataset, ML classifiers based on traditional NLP techniques typically struggle even with the help of over/under sampling techniques like SMOTE. In addition, since proactive events are sometimes subjectively classified as safety related events, a deeper understanding of language is required to detect the nuance in a proactive safety report which can only be achieved through use of attention based large language models such as GPT-4.

### 3.1  Station Condition Record (SCR) Length Distribution

The SCRs ranged in size from 43 to 7998 tokens, with a mean of 584.4 and a median of 368 tokens. Given the small record size, summarization is not required to fit within the context window of modern LLMs like GPT-4 with >8k tokens of context window. The number of output tokens is very small (just the classification of YES/NO, often in JSON format to maximize consistency).

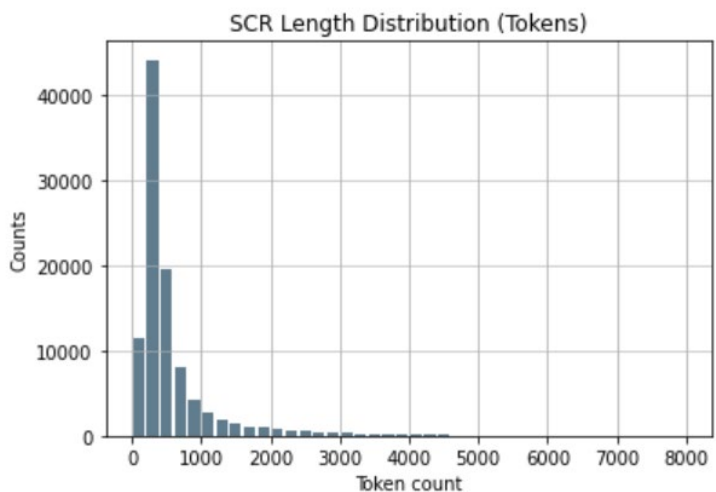

Figure 1 – SCR Length Distribution

### 3.2  Vector Embeddings

To better represent the relationship between SCRs, we first concatenated all the relevant text data, and then computed vector embeddings from this combined text field, using OpenAI's ada2 embeddings endpoint. The resulting embeddings compress the information contained in each SCR (regardless of length) into a vector of 1536 numbers. These 'semantic features' are computed based on the meaning of the text by the transformer encoder architecture of OpenAI's ada2 model. These semantic embeddings allow SCRs to be visualized and compared based on their contents.
t-SNE (t-distributed Stochastic Neighbor Embedding) is an unsupervised dimensionality reduction technique used to visualize high-dimensional data. More similar embeddings are represented close to each other, allowing clusters of similar SCRs to be identified. A t-SNE representation of all 100k SCR embeddings is shown below, in which it can be observed that most safety-related SCRs are grouped in one area of the graph, comprising several distinct sub-clusters. A smaller number of SCRs classified as safety events are also dispersed among non-safety SCRs.

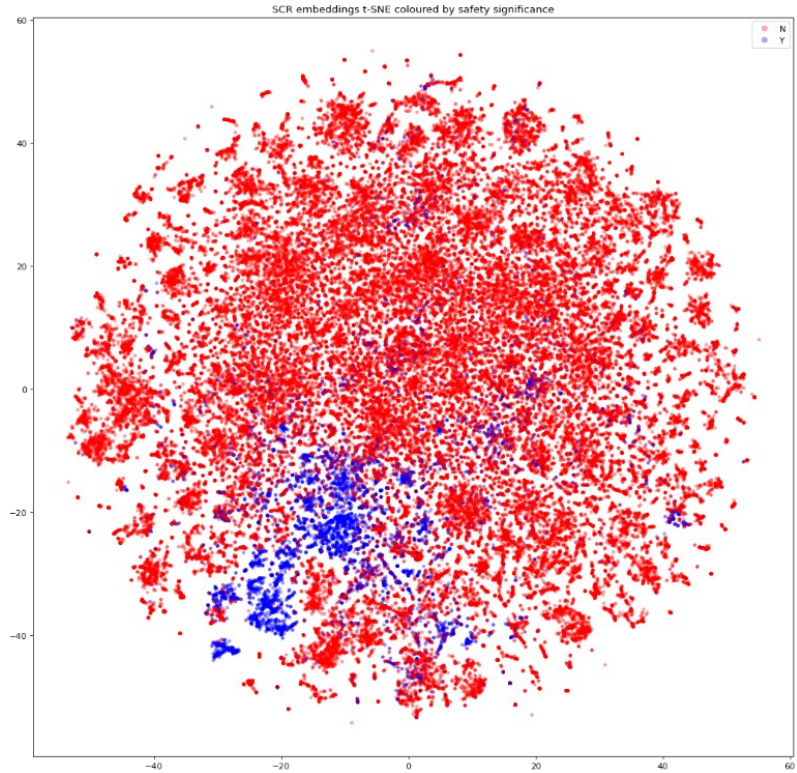

Figure 2 – SCR Embeddings t-SNE Safety Significance

### 3.3  Model Development

As outlined in the introduction, the objective is to employ GPT-4 for this use case. However, given the close grouping observed in the t-SNE analysis during the exploratory data analysis (EDA) phase, it is advisable to initially test a basic classification approach. This will help gauge its performance and establish a baseline for comparison.

### 4.  Evaluation and Results

### 4.1  Nearest Neighbors (k-NN) Classification

This observation of safety-related SCR embeddings clustering together prompted us to experiment with using traditional machine learning techniques to classify SCRs as safety- or non-safety-related. Specifically, k-Nearest Neighbors (k-NN), a supervised machine learning approach, was employed.
A k-NN model was trained on a sample of 10,000 SCRs using k=5, meaning samples were classified based on the 5 SCRs with embeddings closest to their own. The results of this experiment are shown in Figure 4.

Although the overall accuracy of this method was promising, it came at the cost of poor performance on safety records, which are the minority class in this dataset.
Due to the poor recall of this sample on safety-related SCRs, the decision was made to develop a classifier using an LLM specifically OpenAI's GPT models. Ultimately, it is preferred for this model to have a higher False Negative rate than a high False Positive Rate since we want to catch every safety even without missing any, if we have a few false positives that is ok as the humans can screen them out easily.

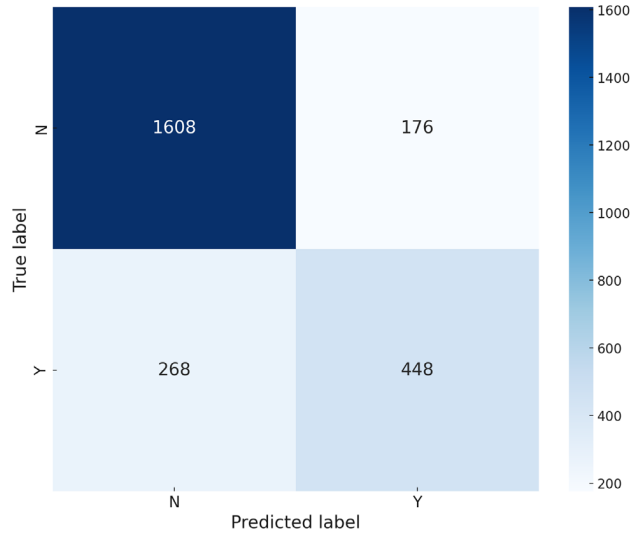

Figure 3 – k-NN Classification Safety Significance

### *4.2 GPT-4 based Classifier*

GPT-4 was the model selected for this problem because the task involves high-level reasoning and complex decision-making based on consideration of context and informed assumptions. In short, GPT-4 is OpenAI's most advanced and capable model, and this task was deemed to be difficult enough to warrant its use. If the results are successful, smaller open-source models (such as Mistral, Phi-2) can be swapped in, potentially after being fine-tuned on a labelled dataset.

4.2.1 Prompt 1 – Basic Prompt

Initial experiments yielded unsatisfactory results. In one early run, GPT-4 was given the following system prompt along with the combined text field for each SCR.

```
You are an AI assistant designed for helping classify safety events.
Events are filed by employees for anything from how well a project went to actual
safety situations that could cause harm.
One element of a safety event, is when someone gets hurt either as a direct result
of work or any other medical event.
Even if someone doesn't get hurt, if there are any unsafe conditions that could
cause harm to workers or put them in unsafe conditions, it is still considered a
safety event.
Please respond in json format: { "safety": "Y/N" }.
```

The results produced by this model were strongly biased toward positive safety predictions. Although the dataset was comprised of over 90% non-safety events, the model predicted almost 70% as safety events. The overall accuracy of the model on all 100,000 SCRs was just 29.3%.

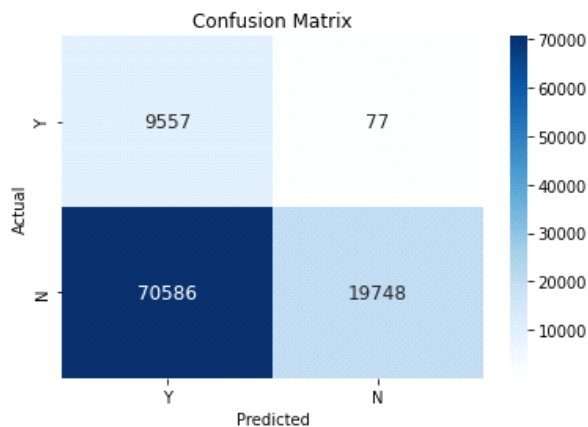

Figure 4 – Confusion Matrix for Prompt 1

Note how many non-safety related SCRs are falsely classified as safety SCRs in this experiment. While this model performed very poorly on non-safety SCRs (recall of 21.9%), it may be noted that it performed quite well on safety SCRs (recall of 99.2%)

4.2.2 <u>Prompt 2 – Safety Culture Adjustments</u>

Overall, we noted that OPG's strong safety culture was biasing the results. There is a tendency for staff to tie almost any and all events back to safety in some way, this is simply the way work gets prioritized based on the safety impact, so even technical problems often have an anecdote of safety mentioned in the writeup.

Improvements were made to the system prompt to be more specific on what criteria constitutes a safety event to improve the results. In consultation with Health and Safety personnel three factors were identified which were used by managers when assessing the safety-relevance of SCRs. These were incorporated into a revised prompt, one version of which is shown below.

```
You are an AI assistant designed to classify reports as either safety events or
non-safety events.
Consider the following factors when making this determination:
1. Does this report describe an event in which someone was injured, a situation in
which someone was ill, or a situation that could have directly caused someone to
be injured?
2. Does this report describe an adverse physical condition with potential to
directly cause an injury or illness?
3. Does this information describe a safety hazard or actions taken to correct a
safety hazard?
Please respond in json format: { "safety": "Y/N" }.
```

These changes improved the model's performance significantly, as shown in the following results based on a trial of 10,000 SCRs.

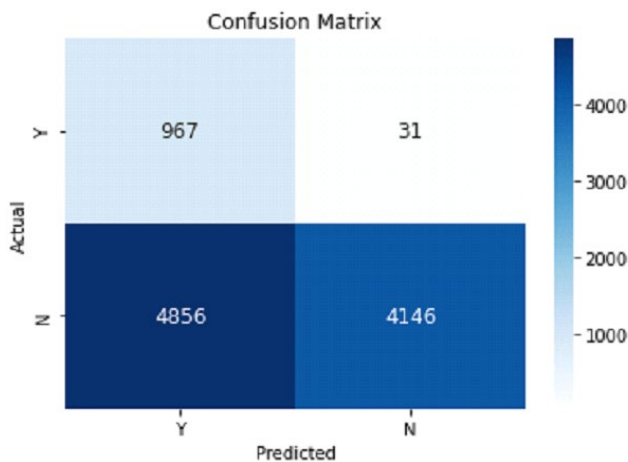

Figure 5 – Confusion Matrix for Prompt 2

With this prompt the model is still biased toward positive classifications, and hence still suffers from a high rate of false-positive classifications. The recall on positive samples is 46.1%, and the overall accuracy is 51.1%, which still leaves much to be desired.

### 4.2.3  Prompt 3 – Corrective Action Adjustments + Potential for Harm

After reviewing some of the false-positive records, it was noted that corrective actions were often recommended using safety as a justification for their priotization even though they didn't necessarily stem from a safety event. So modifications to the system prompt were devised to reduce the likelihood of erroneous positive classification by explaining the corrective action context, and also adding a disclaimer that if the risk/potential for harm is very low, it shouldn't be counted. These changes resulted in the following prompt:

```
You are an AI assistant designed to classify reports as either safety events or
non-safety events.
Consider the following factors when making this determination:
1. Does this report describe an event in which someone was injured, a situation in
which someone was ill, or a situation that could have directly caused someone to
be injured?
2. Does this report describe an adverse physical condition with potential to
directly cause an injury or illness?
3. Does this information describe a safety hazard or actions taken to correct a
safety hazard?
Corrective actions taken do not necessarily indicate an event is safety-related
unless those actions are taken to address direct threats to personnel safety.
In addition, the potential to cause harm may not be sufficient to warrant a
safety-related status if the likelihood of potential harm is very small.
```

```
Please respond in json format: { "safety": "Y/N" }.
```

These modifications produced a marked increase in overall performance and positive recall. This version of the model had an accuracy of 78.9% and positive recall of 89.3%

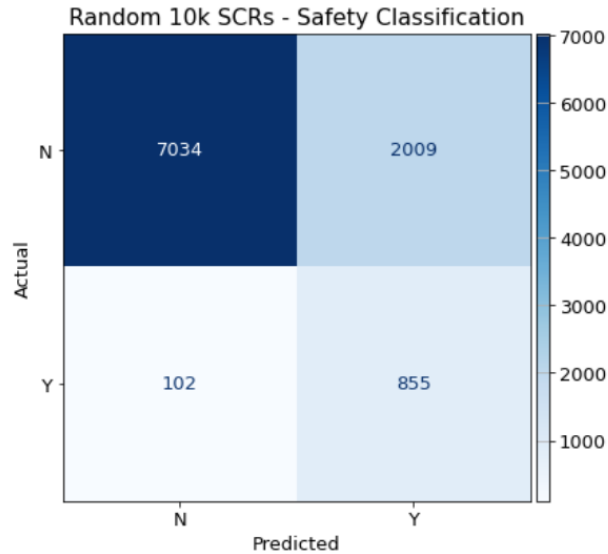

Figure 6 – Confusion Matrix for Prompt 3

### 4.2.4  Prompt 4 – Score Introduction

To address these concerns a final improvement was implemented - the model was instructed to output a numerical score representing a record's safety-relatedness. This gives us greater control over the performance of the model, by selecting the threshold at which an SCR would be classified as a safety event. This change was contained in the following system prompt:

```
In addition, please assign a score of safety-relatedness, as a number from 0 (non-
safety-related) to 1 (strongly safety-related).
Consider the following factors when making this determination:
1. Does this report describe an event in which someone was injured, a situation in
which someone was ill, or a situation that could have directly caused someone to
be injured?
2. Does this report describe an adverse physical condition with potential to
directly cause an injury or illness?
3. Does this information describe a safety hazard or actions taken to correct a
safety hazard?
Corrective actions taken do not necessarily indicate an event is safety-related
unless those actions are taken to address direct threats to personnel safety.
In addition, the potential to cause harm may not be sufficient to warrant a
safety-related status if the likelihood of potential harm is very small.
With x denoting the safety-relatedness score between 0 and 1, please respond in
json format: {"safety score": x, "safety": "Y/N"}.
```

Here is a sample distribution (across 5k SCRs) of the safety scores produced by GPT-4 when given this prompt:

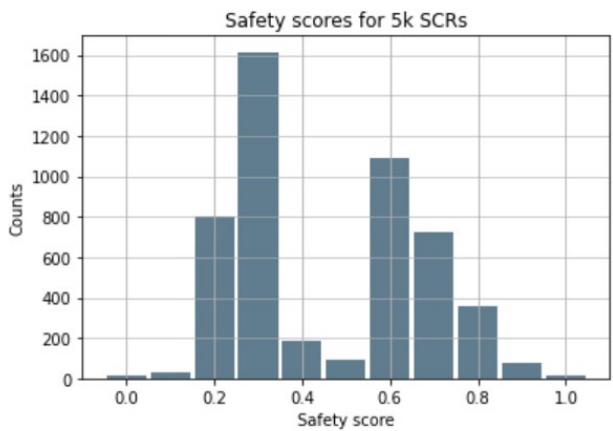

Figure 7 – Safety Scores for 5000 SCRs via Prompt 4

This allowed us to select different safety classification thresholds and compare results. For example, whether the threshold for a safety='Y' classification is set at 0.5 or 0.6.

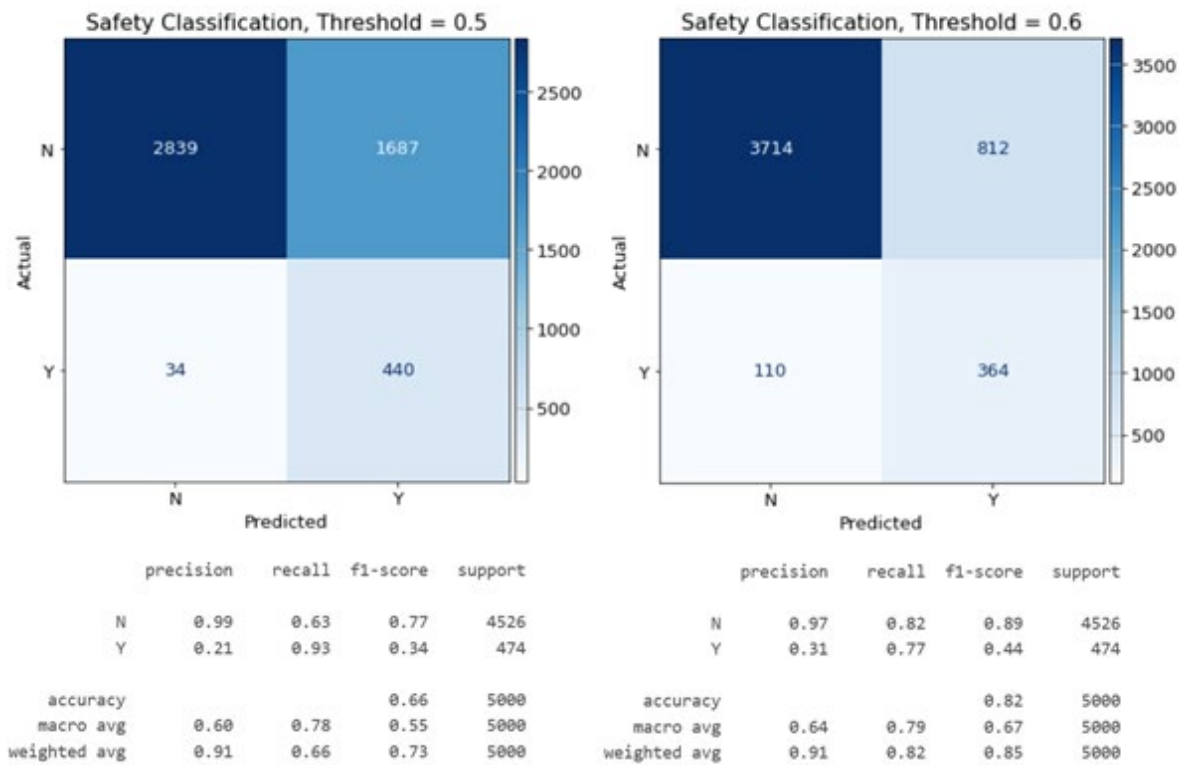

Figure 8 – Confusion Matrix, Precision, Recall, F1 Score for Prompt 4

### 4.2.5 Prompt 5 – Score control + Final Improvements

One drawback of this specific version of the prompt was that the model always output safety scores as multiples of 0.1 (e.g. 0.1 or 0.4, never 0.27). This limited our control over the classification threshold, as it could only be set at a small number of points.
This was improved with the following prompt, which also includes a few miscellaneous improvements mainly the inclusion of JHSC as a trigger to guarantee a safety detection:

```
You are an AI assistant designed to classify reports as either safety events or
non-safety events.
In addition, please assign a score of safety-relatedness, as a number from 0 (non-
safety-related) to 100 (strongly safety-related).  Please give granular scores
including numbers that are not multiples of 5 or 10 (e.g. 53, 14, 62).
Consider the following factors:
1. Does this report discuss an injury or potential injury, a situation in which
someone was ill, or a situation that could have directly caused someone to be
injured?
2. Does this report describe an adverse physical condition or behavior with
potential to directly cause an injury or illness?
3. Does this information describe a safety hazard or actions taken to correct a
safety hazard, or safety conditions that are non-compliant with policies and
procedures?
Any record that mentions "Joint Health & Safety Committee" or "JHSC" is a safety
event.
Corrective actions taken do not necessarily indicate an event is safety-related
unless those actions are taken to address direct threats to personnel safety.
In addition, the potential to cause harm may not be sufficient to warrant a
safety-related status if the likelihood of potential harm is very small.
With x denoting the safety-relatedness score between 0 and 100, please respond in
json format: {"safety score": x, "safety": "Y/N"}.
```

With this prompt, a more granular distribution of safety scores between 0 and 100 was produced, as shown below.

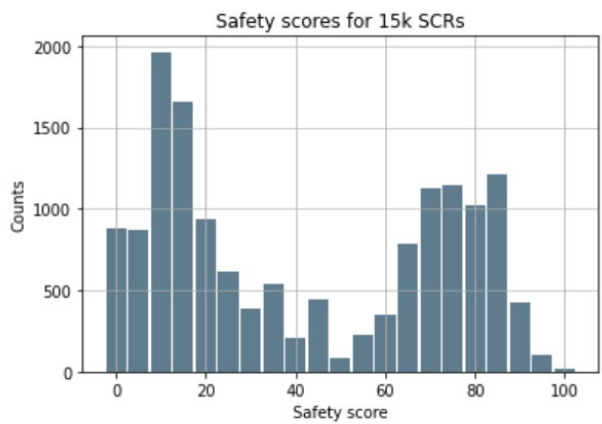

Figure 9 – Safety Scores for 15 000 SCRs via Prompt 5

This model produced the following performance at selected classification thresholds.

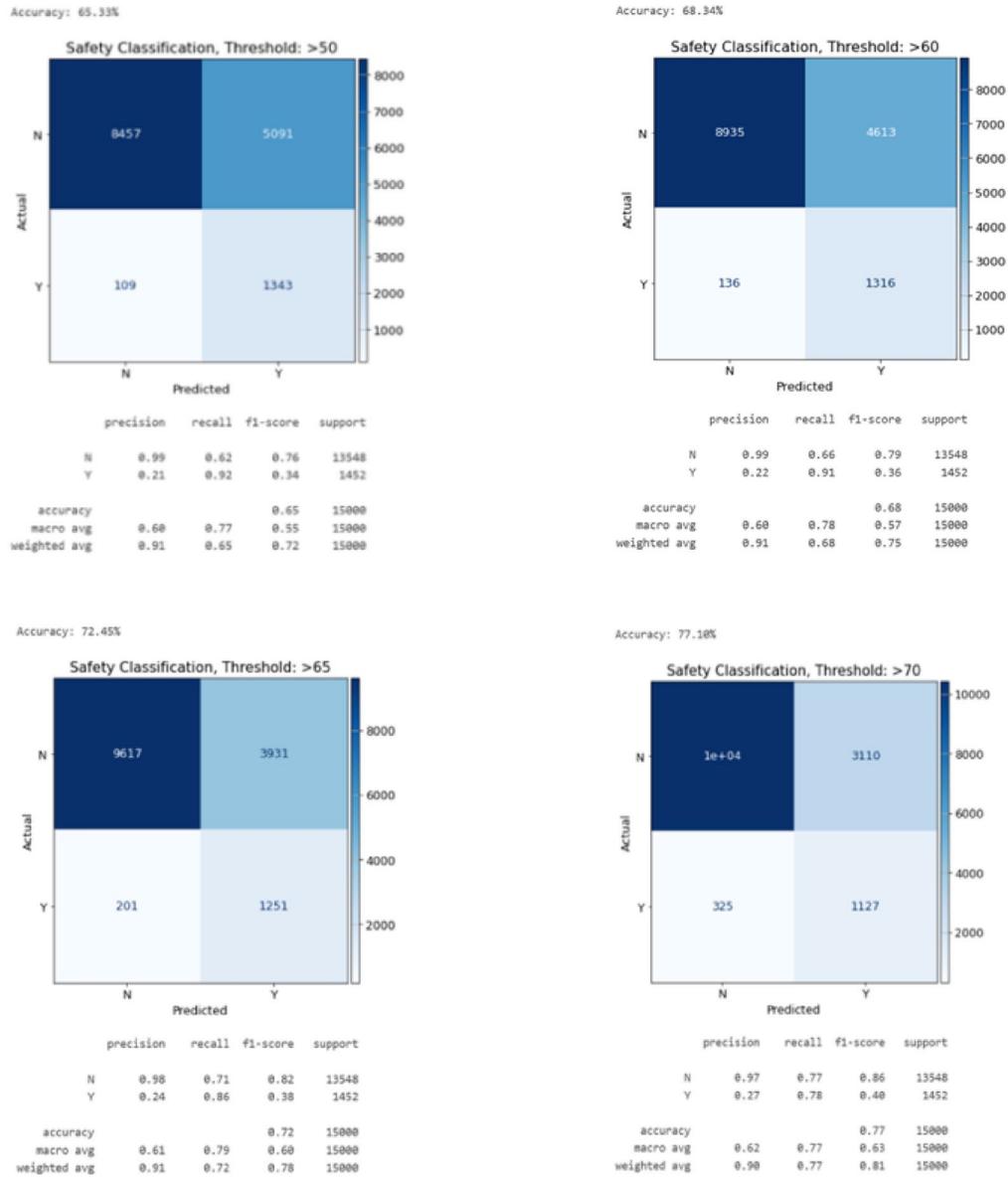

Figure 10 - Confusion Matrix, Precision, Recall, F1 Score for Prompt 5

Based on a classification threshold of 70, this model achieved overall accuracy of 77.1%, while still maintaining a positive recall of 78% and a negative recall of 77%. These results (regardless of which specific threshold is used) represent the best performance achieved on this problem to date, which promises to translate into a useful software innovation with the potential to improve identification of safety events at OPG.

## 5. Conclusions and Future Work

In conclusion, we have developed a model that performs sufficiently well for the Health and Safety department to consider its deployment to enhance efficiency. While the model is not without flaws, its current level of performance justifies a pilot deployment due to the significant potential gains in worker efficiency and safety. Looking ahead, several improvements are suggested: Firstly, transitioning to GPT-4 Turbo, which offers a larger context window allowing for handling extensive SCRs without the need for summarization that might compromise quality or context. Additionally, this model improves instruction following capabilities, reducing the risk of format errors in safety scoring outputs—currently managed with retry logic that increases runtime. Secondly, integrating GPT technology directly into the data intake process could address issues with limited or vague SCR data. A proposed chatbot could interactively refine data input, running the AI classifier during record entry and prompting for more details when confidence in the record's detail is low, thereby improving prediction quality. Lastly, considering an ensemble model combining the XGBOOST Classifier, which, despite its lower overall performance compared to ChatGPT, has shown high recall, could potentially enhance results further. This hybrid approach may leverage the strengths of both models to achieve superior outcomes in safety event identification.

## 6. Acknowledgements

This research was funded by The Natural Sciences and Engineering Research Council of Canada (NSERC) and the Canadian Nuclear Safety Commission (CNSC).